\documentclass[10pt,twocolumn,letterpaper]{article}

\usepackage[pagenumbers]{cvpr} 

\usepackage{times}  
\usepackage{helvet}  
\usepackage{courier}  
\usepackage[hyphens]{url}  
\usepackage{graphicx} 
\usepackage{natbib}  
\usepackage{caption} 

\usepackage{algorithm}
\usepackage{algorithmic}
\usepackage{enumitem}
\usepackage[utf8]{inputenc} 
\usepackage{booktabs}       
\usepackage{nicefrac}       
\usepackage{microtype}      
\usepackage{xcolor}         
\usepackage{multirow}
\usepackage{amsmath}
\usepackage{caption}
\usepackage{enumitem} 
\usepackage{amssymb}    
\usepackage{amsfonts}   
\usepackage{graphicx}   
\usepackage{bbm}        
\usepackage{mathtools} 
\usepackage{makecell}

\usepackage{adjustbox}

\usepackage{newfloat}
\usepackage{listings}

\definecolor{cvprblue}{rgb}{0.21,0.49,0.74}
\usepackage[pagebackref,breaklinks,colorlinks,allcolors=cvprblue]{hyperref}

\def\paperID{*****} 
\def\confName{CVPR}
\def\confYear{2026}

\title{Vector-Quantized Soft Label Compression for Dataset Distillation}

\author{Ali Abbasi\\
Vanderbilt University\\
{\tt\small ali.abbasi@vanderbilt.edu}
\and
Ashkan Shahbazi\\
Vanderbilt University\\
{\tt\small ashkan.shahbazi@vanderbilt.edu}
\and
Hamed Pirsiavash\\
University of California, Davis\\
{\tt\small hpirsiav@ucdavis.edu}
\and
Soheil Kolouri\\
Vanderbilt University\\
{\tt\small soheil.kolouri@vanderbilt.edu}
}

\begin{document}
\maketitle

\begin{abstract}
 Dataset distillation is an emerging technique for reducing the computational and storage costs of training machine learning models by synthesizing a small, informative subset of data that captures the essential characteristics of a much larger dataset. Recent methods pair synthetic samples and their augmentations with soft labels from a teacher model, enabling student models to generalize effectively despite the small size of the distilled dataset. While soft labels are critical for effective distillation, the storage and communication overhead they incur—especially when accounting for augmentations—is often overlooked. In practice, each distilled sample is associated with multiple soft labels, making them the dominant contributor to storage costs, particularly in large-class settings such as ImageNet-1K. In this paper, we present a rigorous analysis of bit requirements across dataset distillation frameworks, quantifying the storage demands of both distilled samples and their soft labels. To address the overhead, we introduce a vector-quantized autoencoder (VQAE) for compressing soft labels, achieving substantial compression while preserving the effectiveness of the distilled data. We validate our method on both vision and language distillation benchmarks. On ImageNet-1K, our proposed VQAE achieves 30--40$\times$ additional compression over RDED, LPLD, SRE2L, and CDA baselines while retaining over $90\%$ of their original performance.
\end{abstract}

\section{Introduction}

Modern machine learning models owe much of their success to access to large-scale, high-quality datasets. As model capacity continues to grow, so does the need for massive data, making data collection, storage, and training increasingly expensive. These challenges have sparked interest in dataset distillation \cite{wang2018dataset, zhao2021dataset}, a technique that aims to synthesize a small, informative set of training examples from a large dataset, such that models trained on the distilled data perform comparably to those trained on the full dataset. By replacing millions of real samples with a compact synthetic core, dataset distillation offers a promising path toward reducing the computational and storage overhead of model training, enabling scalable learning in resource-constrained settings \cite{cazenavette2022dataset,cui2023scaling,yin2023squeeze}.

The dataset distillation problem is traditionally formulated as a bi-level optimization task. In this setting, the inner loop optimizes the parameters of a model trained on the distilled dataset, and the outer loop updates the synthetic dataset itself to minimize the model’s loss on the validation set of the original dataset \cite{wang2018dataset}. Formally, this requires unrolling the entire training process of the model and backpropagating gradients through each optimization step, which is computationally expensive and memory-intensive. To alleviate this, a variety of strategies have been proposed, including gradient matching \cite{zhao2021dataset}, trajectory matching \cite{cazenavette2022dataset}, kernel-based methods \cite{nguyen2021dataset}, and distribution matching \cite{zhao2023dataset}. While these approaches have shown success on small-scale datasets such as CIFAR-10 and Tiny-ImageNet, they often struggle to scale to larger datasets and models, such as ImageNet-1K or transformer-based architectures, due to the prohibitive cost of backpropagation through long training trajectories.


\begin{figure*}[t!]
    \centering
    \includegraphics[width=.95\linewidth]{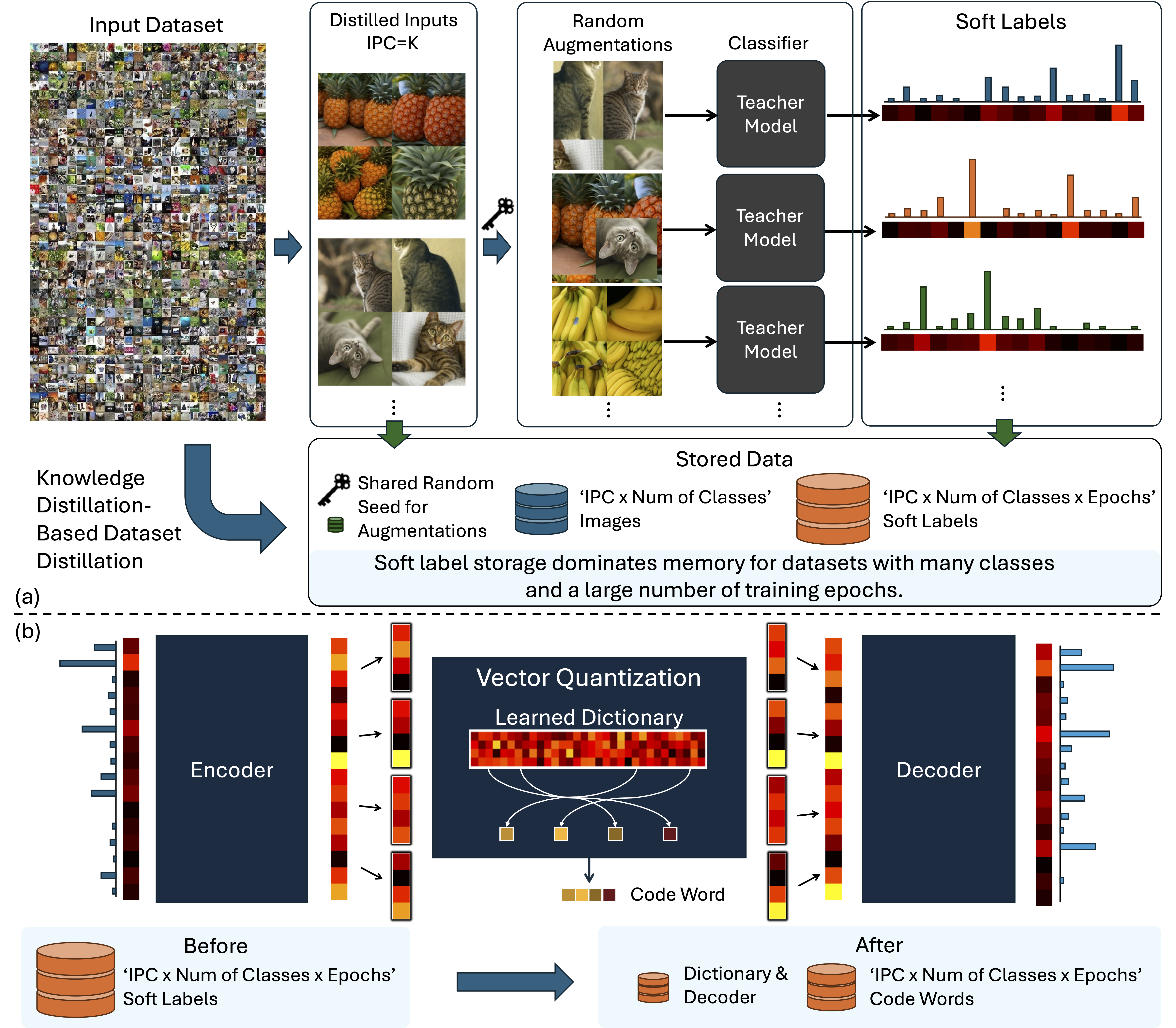}
    \caption{(a) We begin with a standard dataset distillation pipeline, where a small number of distilled inputs per class (IPC) are passed through a teacher model to produce soft labels across multiple augmentations. These soft labels, although highly informative, impose a significant memory and communication burden, especially for large numbers of classes and training epochs.
 (b) To mitigate this overhead, we introduce a vector-quantized autoencoder (VQ-AE) that encodes soft labels into compact latent representations segmented and quantized using a learned dictionary. Only the code indices, dictionary, and decoder need to be stored and transmitted. At distillation time, soft labels are reconstructed from these compact codes and used to train student models via KL divergence. This approach dramatically reduces storage while preserving the fidelity of the teacher’s knowledge.}
    \label{fig:teaser}
\end{figure*}

To overcome the scalability limitations of bi-level optimization, recent methods have shifted toward a decoupled training paradigm in which a pretrained teacher model guides the synthesis of distilled data. Instead of differentiating through the training process, recent methods leverage a fixed teacher model (pretrained on the full dataset) to guide the synthesis of training data and generate supervision in the form of soft labels, i.e., class probability distributions predicted by the teacher for each synthetic sample. The student model is then trained on the distilled dataset via knowledge distillation \cite{hinton2015distilling,yin2023squeeze,yin2024dataset}. These approaches have successfully scaled dataset distillation to large benchmarks like ImageNet-1K, achieving state-of-the-art performance \cite{yin2024dataset,sun2024diversity,xiao2024are,qin2024a}. The synthetic data generation strategies vary widely, including matching batch normalization statistics or feature distributions \cite{yin2023squeeze}, and constructing realistic collages from discriminative image patches \cite{sun2024diversity}. Regardless of the method used to compress the dataset, in all these methods, the teacher provides soft labels that are stored and used during the student training phase (see Figure \ref{fig:teaser} (a)). These soft labels, often computed over hundreds of augmentations per image, serve as rich, distributional supervision signals and are widely credited as a key factor behind the success of modern dataset distillation techniques \cite{qin2024a}.

Interestingly, recent studies have demonstrated that the structure of the distilled images may be less critical than previously thought, and that the use of soft labels is the dominant contributor to performance \cite{xiao2024are,qin2024a}. Qin et al. \cite{qin2024a} show that simply pairing randomly selected images from the original dataset with soft labels generated by a pretrained teacher, without any image synthesis or optimization, can match or even outperform several state-of-the-art dataset distillation methods. Their findings suggest that the knowledge encoded in soft labels, especially when generated across multiple augmentations, plays a disproportionately large role in enabling generalization from compact datasets.

This shift in emphasis toward soft-label supervision brings to light a largely overlooked cost in modern dataset distillation: the storage overhead of soft labels. In many recent methods, including SRe2L \cite{yin2023squeeze}, CDA \cite{yin2024dataset}, and RDED \cite{sun2024diversity}, and the growing body of diffusion-based dataset distillation frameworks \cite{su2024d,chen2025influenceguided}, soft labels are precomputed for every synthetic image (and each of its augmentations), and stored at 16-bit or 32-bit floating-point precision. For datasets with a large number of classes, e.g., 1,000 for ImageNet-1K, 10,450 for ImageNet-21K, or 50,000–150,000 for token-level tasks in NLP, the cumulative storage requirement for soft labels can far exceed that of the image/text data itself.  The significance and cost of soft labels for dataset distillation have only recently been highlighted \cite{qin2024a,xiao2024are} and remain relatively underexplored, which limits the application of dataset distillation at scale.

In this paper, we focus specifically on the storage cost of soft labels in teacher-guided dataset distillation. After quantifying the substantial overhead introduced by storing teacher outputs, particularly in large-scale settings involving images or token-level language tasks, we propose a simple and effective solution: a vector-quantized autoencoder for compressing soft labels (Figure \ref{fig:teaser} (b)). Our method learns a discrete codebook of label prototypes and encodes teacher outputs via compact latent indices, significantly reducing the memory footprint while preserving label fidelity. Our label compression is orthogonal to how the input data is distilled, and can be integrated seamlessly with existing input condensation methods such as SRe2L \cite{yin2023squeeze}, CDA \cite{yin2024dataset}, RDED \cite{sun2024diversity}, and LPLD \cite{xiao2024are}.

One application of our method is in LLM distillation. Typically, a large model is trained by company A and then distilled into a smaller model by company B. This setup requires company B to access and run the large teacher model, an expensive process that also depends on company A’s willingness to share the model. An alternative is for company A to precompute and share soft labels, but these are prohibitively large (e.g., $>$50K values per token). Our method offers an efficient solution by compressing soft labels before transfer, allowing company B to perform token-level distillation without accessing the large model or significant compute. Our results in this setting are promising.

\section{Related Work}

\textbf{Dataset Distillation and Soft Labels.}
Dataset distillation aims to generate a small synthetic dataset such that models trained on it achieve performance comparable to training on the full dataset~\cite{wang2018dataset}. Existing methods fall into four broad categories: (1) \emph{Bi-level optimization}, which formulates the task as a nested optimization over model weights and synthetic data~\cite{zhao2021dataset,cazenavette2022dataset,kim2022dataset,zhang2023accelerating}; (2) \emph{Uni-level relaxations}, including NTK-based methods~\cite{nguyen2021dataset,loo2023dataset} and decoupled training frameworks such as SRe2L~\cite{yin2023squeeze} and CDA~\cite{yin2024dataset}; (3) \emph{Coreset selection}, which selects real data subsets using bi-level optimization~\cite{borsos2020coresets,toneva2018empirical,paul2021deep}; and (4) \emph{Generative priors}, which use pretrained generative models (e.g., diffusion or GAN models) to regularize the synthesis process~\cite{cazenavette2023generalizing,sun2024diversity,abbasi2024diffusion,abbasi2024one,su2024d,gu2024efficient,chen2025influenceguided}.

A common thread in recent scalable approaches is the use of a pretrained \emph{teacher model} to guide both input synthesis and label generation. Methods such as SRe2L~\cite{yin2023squeeze}, CDA~\cite{yin2024dataset}, LPLD~\cite{xiao2024are}, and RDED~\cite{sun2024diversity} generate soft labels by forwarding synthetic or real crops through a teacher network. Qin et al.~\cite{qin2024a} show that performance gains in these methods often stem from the use of soft labels rather than the particular image synthesis strategy. This growing reliance on soft-label supervision highlights a critical but underexplored bottleneck: the memory cost of storing per-sample, per-augmentation label distributions especially in large-class or token-level tasks.

\textbf{Label Compression in Knowledge Distillation.}
The problem of compressing teacher outputs has been explored in the broader knowledge distillation (KD) literature~\cite{hinton2015distilling}. The overhead of soft-label storage has received increasing attention in recent works~\cite{yun2021re,shen2022fast,shen2023ferkd}. Most relevant to our framework are FKD~\cite{shen2022fast} and FerKD~\cite{shen2023ferkd}, which propose storing region-level soft labels offline to eliminate the runtime cost of teacher forward passes. FKD introduces a mechanism for precomputing and reusing region-specific soft labels across training epochs, and explores several compression strategies, such as hardening, label smoothing, and top-$K$ marginal smoothing, to reduce memory usage. FerKD extends this idea by introducing a partial label adaptation mechanism that selectively replaces unreliable soft labels, typically from background or ambiguous regions, with smoothed hard labels, while retaining soft labels for informative regions. Although these methods reduce redundancy and improve supervision quality, they do not directly address lossy compression of soft labels via learned representations, which is the focus of our work. In the context of dataset distillation, recently, Xiao \& He \cite{xiao2024are} proposed to perform random batch pruning to reduce the overhead of soft-label storage. In a concurrent work, Yuan et al. \cite{yuan2025score} utilize robust PCA for compressing the soft-labels. Inspired by the recent success of Vector Quantized Auto-Encoders (VQAEs), we focus on learnable lossy compression techniques for soft-label compression.

\textbf{Vector Quantization for Representation Compression.}
Vector quantization has been widely used for compressing continuous representations via discrete codebooks, most notably in VQ-VAE~\cite{van2017neural} and its derivatives~\cite{esser2021taming}. We adopt this paradigm for compressing soft labels, learning a discrete embedding space for teacher outputs. To our knowledge, this is the first application of vector quantization for scalable soft-label compression in dataset distillation. We will provide comprehensive comparisons and ablation studies to show the effectiveness of VQ for soft-label compression.

\section{Method}

We propose a two-stage framework for compressing and distilling soft labels in knowledge distillation: a \emph{caching stage}, where the teacher's output probabilities (soft-labels) are encoded via a vector-quantized autoencoder (VQAE), and a \emph{distillation stage}, where soft labels are reconstructed from codebook indices. This approach dramatically reduces the memory and communication overhead of soft label storage while preserving their informative signal.

\subsection{Preliminaries and Notation}

Let $\mathcal{X} \subset \mathbb{R}^d$ denote the input space, and let $c$ be the number of output classes. We consider a teacher model $T(\cdot; \theta_T): \mathcal{X} \rightarrow \mathbb{R}^c$ and a student model $S(\cdot; \theta_S): \mathcal{X} \rightarrow \mathbb{R}^c$. Given an input $x \in \mathcal{X}$, the teacher produces logits $\mathbf{z} = T(x) \in \mathbb{R}^c$, which we convert to a soft label distribution via softmax: $\mathbf{y} = \text{softmax}(\frac{\mathbf{z}}{\tau}) \in \Delta^{c-1}$, where $\Delta^{c-1}$ denotes the c-dimensional simplex, and $\tau$ is the temperature. 

One can store all inputs and their soft labels and then train a student by knowledge distillation using:
\begin{equation}
\mathcal{L}_{\text{KD}} = \text{KL}\left({\mathbf{y}} \bigg\|\, \text{softmax}\left(\frac{S(x)}{\tau}\right) \right), \label{eq: kl_loss}    
\end{equation}

\noindent where $S(.)$ is the logits of the student model, and $KL(P||Q)=\sum_{x\in\mathcal{X}} P(x)log(P(x)/Q(x))$ is the Kullback-Leibler divergence between two probability distributions. However, this approach requires storing and communicating all soft labels, which is costly. In the following, we introduce a vector-quantized autoencoder to compress the knowledge in the soft labels.

\subsection{Soft Label Compression via Vector-Quantized Autoencoding}

To reduce storage cost, we compress $\mathbf{y}$ using a vector-quantized autoencoder. After experimenting with several architectures, we found that a simple linear encoder-decoder is sufficient. Specifically, we first project $\mathbf{y} \in \mathbb{R}^c$ into a latent space using a linear encoder matrix $P \in \mathbb{R}^{c \times d_h}$:

\begin{equation}
\mathbf{h} = \mathbf{y} P \in \mathbb{R}^{d_h}. \label{eq:encode}
\end{equation}

We then partition the latent vector $\mathbf{h}$ into $m$ equal-sized segments, each of dimension $d_c = d_h / m$:

\begin{equation}
\mathbf{h} = [\mathbf{h}^{(1)}, \dots, \mathbf{h}^{(m)}], \quad \mathbf{h}^{(i)} \in \mathbb{R}^{d_c}. \label{eq:latent_formation}
\end{equation}

Each segment $\mathbf{h}^{(i)}$ is quantized using a shared codebook $\mathcal{\mu} = \{\boldsymbol{\mu}_1, \dots, \boldsymbol{\mu}_k\} \subset \mathbb{R}^{d_c}$. We assign each segment to its nearest code vector:
%
\begin{equation}
q^{(i)} = \arg \min_j \| \mathbf{h}^{(i)} - \boldsymbol{\mu}_j \|_2^2, \quad \hat{\mathbf{h}}^{(i)} = \text{sg}[\boldsymbol{\mu}_{q^{(i)}}], \label{eq:nearest_code}
\end{equation}

where $\text{sg}[\cdot]$ denotes the stop-gradient operator. The quantized latent is formed by concatenating the quantized segments:

\begin{equation}
\hat{\mathbf{h}} = [\hat{\mathbf{h}}^{(1)}, \dots, \hat{\mathbf{h}}^{(m)}] \in \mathbb{R}^{d_h}. \label{eq:quantized_formation}
\end{equation}

Finally, we reconstruct the original vector using a linear decoder matrix $D \in \mathbb{R}^{d_h \times c}$:

\begin{equation}
\hat{\mathbf{y}} = \hat{\mathbf{h}} D \in \mathbb{R}^c. \label{eq:reconstruction}
\end{equation}

\subsection{Caching Loss and Optimization}

The compression module is trained by minimizing the reconstruction error between the original probabilities and their reconstruction, along with the standard VQ-AE and codebook losses:




\begin{align}
\mathcal{L}_{\text{cache}} &=
\alpha \underbrace{\sum_{i=1}^m \big( \| \text{sg}[\mathbf{h}^{(i)}] - \boldsymbol{\mu}_{q^{(i)}} \|_2^2
+ \beta \| \mathbf{h}^{(i)} - \text{sg}[\boldsymbol{\mu}_{q^{(i)}}] \|_2^2 \big)}_{\mathcal{L}_{\text{VQ}}} \notag\\
&\quad + \underbrace{\| \hat{\mathbf{y}} - \mathbf{y} \|_2^2}_{\mathcal{L}_{\text{rec}}}
\label{eq:objective}
\end{align}


\noindent where $\alpha$ and $\beta$ are hyperparameters balancing three loss terms. The trainable components include the encoder matrix $P$, decoder matrix $D$, and the codebook $\mathcal{\mu}$. Once trained, all components are frozen and used for soft label compression and reconstruction.

{\bf Compression ratio:} To store or transmit the soft labels, we only need to save the $m$ quantized indices per soft label (one per segment), along with the codebook of size $k d_c$ and the decoder projection matrix of size $d_h \times c$. Thus, the total storage cost for $n$ images with $a$ augmentations each is:
\[
a n m + k d_c + c d_h.
\]
Assuming a large number of categories ($c \gg m$), this is significantly smaller than storing the full soft labels, which requires $a n c$ values. Consequently, the compression ratio for soft labels alone is:
\begin{equation}
    \frac{a n c}{a n m + k d_c + c d_h}. \label{eq:comp_ratio}
\end{equation}

In the above statement, we count the number of numbers only, but in the experiments, we count the actual number of bits and consider JPEG compression to report the actual size on disk in MBs.

\subsection{Soft Label Reconstruction and Distillation}




At distillation time, we only transmit the quantized code indices $\{q^{(1)}, \dots, q^{(m)}\}$. The student retrieves the corresponding codebook vectors and reconstructs
\[
\hat{\mathbf{y}} = [\boldsymbol{\mu}_{q^{(1)}}, \dots, \boldsymbol{\mu}_{q^{(m)}}] D \in \mathbb{R}^c.
\]
Since the linear decoder does not guarantee $\hat{\mathbf{y}} \in \Delta^{c-1}$, we renormalize it to obtain a valid probability vector:
\begin{equation}
\tilde{\mathbf{y}}_j
= \frac{\max(\hat{\mathbf{y}}_j,\epsilon)}{\sum_{\ell=1}^{c}\max(\hat{\mathbf{y}}_\ell,\epsilon)},
\qquad j=1,\dots,c,
\label{eq:renorm}
\end{equation}
where $\epsilon>0$ is a small constant.
Finally, the student is trained using a softened KL divergence loss:
\begin{equation}
\mathcal{L}_{\text{KD}} = \text{KL}\left(\tilde{\mathbf{y}} \bigg\|\, \text{softmax}\left(\frac{S(x)}{\tau}\right) \right).
\label{eq:kd_loss_renorm}
\end{equation}

\section{Experiments}

\raggedbottom     
\begin{table}[!htbp]
\centering
\setlength{\tabcolsep}{2pt}
\renewcommand{\arraystretch}{1.0} 
\resizebox{\columnwidth}{!}{%
\normalsize
\begin{tabular}{c l cccc}
\toprule
 & \textbf{Method}
 & \textbf{IPC 10}
 & \textbf{IPC 20}
 & \textbf{IPC 50}
 & \textbf{IPC 100} \\
\midrule
\multirow{4}{*}{\rotatebox[origin=c]{90}{\textbf{Comp. 1×}}}
 & RDED  & $39.3{\color{gray}\scriptscriptstyle\pm0.5}$ & $48.3{\color{gray}\scriptscriptstyle\pm0.2}$ & $55.0{\color{gray}\scriptscriptstyle\pm0.1}$ & $58.6{\color{gray}\scriptscriptstyle\pm0.3}$ \\
 & LPLD  & $36.6{\color{gray}\scriptscriptstyle\pm0.8}$ & $47.0{\color{gray}\scriptscriptstyle\pm0.1}$ & $54.8{\color{gray}\scriptscriptstyle\pm0.4}$ & $58.5{\color{gray}\scriptscriptstyle\pm0.2}$ \\
 & SRE2L & $34.2{\color{gray}\scriptscriptstyle\pm0.5}$ & $42.8{\color{gray}\scriptscriptstyle\pm0.4}$ & $50.4{\color{gray}\scriptscriptstyle\pm0.1}$ & $54.9{\color{gray}\scriptscriptstyle\pm0.2}$ \\
  & CDA   & $33.7{\color{gray}\scriptscriptstyle\pm0.2}$ & $43.8{\color{gray}\scriptscriptstyle\pm0.3}$ & $52.1{\color{gray}\scriptscriptstyle\pm0.8}$ & $56.6{\color{gray}\scriptscriptstyle\pm0.2}$ \\
 \midrule[1pt]
\multirow{9}{*}{\rotatebox[origin=c]{90}{\textbf{Compression 10×}}}
 & RDED + LPLD & $39.1{\color{gray}\scriptscriptstyle \phantom{0.000}}$ & $48.1{\color{gray}\scriptscriptstyle\phantom{0.000}}$ & $54.3{\color{gray}\scriptscriptstyle\phantom{0.000}}$ & $57.8{\color{gray}\scriptscriptstyle\phantom{0.000}}$ \\
 & RDED + Ours & {\boldmath$39.7{\color{gray}\scriptscriptstyle\pm0.4}$} & {\boldmath$48.5{\color{gray}\scriptscriptstyle\pm0.3}$} & {\boldmath$55.0{\color{gray}\scriptscriptstyle\pm0.2}$} & $58.6{\color{gray}\scriptscriptstyle\pm0.2}$ \\
 \cmidrule(lr){2-6}
 & LPLD + LPLD & $32.7{\color{gray}\scriptscriptstyle\pm0.6}$ & $44.7{\color{gray}\scriptscriptstyle\pm0.4}$ & $54.4{\color{gray}\scriptscriptstyle\pm0.2}$ & {\boldmath$58.8{\color{gray}\scriptscriptstyle\pm0.1}$} \\
 & LPLD + Ours & $36.7{\color{gray}\scriptscriptstyle\pm0.4}$ & $47.2{\color{gray}\scriptscriptstyle\pm0.6}$ & $54.6{\color{gray}\scriptscriptstyle\pm0.3}$ & $58.5{\color{gray}\scriptscriptstyle\pm0.4}$ \\
 \cmidrule(lr){2-6}
 & SRE2L + LPLD & $18.9{\color{gray}\scriptscriptstyle\phantom{0.000}}$ & $31.1{\color{gray}\scriptscriptstyle\phantom{0.000}}$ & $44.1{\color{gray}\scriptscriptstyle\phantom{0.000}}$ & $51.1{\color{gray}\scriptscriptstyle\phantom{0.000}}$ \\
 & SRE2L + Ours & $33.2{\color{gray}\scriptscriptstyle\pm0.3}$ & $42.6{\color{gray}\scriptscriptstyle\pm0.3}$ & $50.3{\color{gray}\scriptscriptstyle\pm0.3}$ & $55.4{\color{gray}\scriptscriptstyle\pm0.3}$ \\
 \cmidrule(lr){2-6}
 & CDA + LPLD & $28.4{\color{gray}\scriptscriptstyle\phantom{0.000}}$ & $39.7{\color{gray}\scriptscriptstyle\phantom{0.000}}$ & $50.3{\color{gray}\scriptscriptstyle\phantom{0.000}}$ & $55.1{\color{gray}\scriptscriptstyle\phantom{0.000}}$ \\
 & CDA + Ours & $34.0{\color{gray}\scriptscriptstyle\pm0.1}$ & $43.5{\color{gray}\scriptscriptstyle\pm0.3}$ & $52.1{\color{gray}\scriptscriptstyle\pm0.4}$ & $56.7{\color{gray}\scriptscriptstyle\pm0.2}$ \\
\midrule[1pt]
\multirow{9}{*}{\rotatebox[origin=c]{90}{\textbf{Compression 20×}}}
 & RDED + LPLD & $35.7{\color{gray}\scriptscriptstyle\phantom{0.000}}$ & $44.3{\color{gray}\scriptscriptstyle\phantom{0.000}}$ & $52.7{\color{gray}\scriptscriptstyle\phantom{0.000}}$ & $57.1{\color{gray}\scriptscriptstyle\phantom{0.000}}$ \\
 & RDED + Ours & {\boldmath$37.6{\color{gray}\scriptscriptstyle\pm0.3}$} & {\boldmath$47.2{\color{gray}\scriptscriptstyle\pm0.2}$} & {\boldmath$54.0{\color{gray}\scriptscriptstyle\pm0.1}$} & {\boldmath$57.6{\color{gray}\scriptscriptstyle\pm0.1}$} \\
 \cmidrule(lr){2-6}
 & LPLD + LPLD & $28.6{\color{gray}\scriptscriptstyle\pm0.4}$ & $41.0{\color{gray}\scriptscriptstyle\pm0.3}$ & $51.8{\color{gray}\scriptscriptstyle\pm0.2}$ & $57.4{\color{gray}\scriptscriptstyle\pm0.9}$ \\
 & LPLD + Ours & $35.0{\color{gray}\scriptscriptstyle\pm0.5}$ & $46.3{\color{gray}\scriptscriptstyle\pm0.2}$ & $53.5{\color{gray}\scriptscriptstyle\pm0.3}$ & $57.5{\color{gray}\scriptscriptstyle\pm0.1}$ \\
 \cmidrule(lr){2-6}
 & SRE2L + LPLD & $16.0{\color{gray}\scriptscriptstyle\phantom{0.000}}$ & $29.2{\color{gray}\scriptscriptstyle\phantom{0.000}}$ & $41.5{\color{gray}\scriptscriptstyle\phantom{0.000}}$ & $49.5{\color{gray}\scriptscriptstyle\phantom{0.000}}$ \\
 & SRE2L + Ours & $31.7{\color{gray}\scriptscriptstyle\pm0.1}$ & $41.5{\color{gray}\scriptscriptstyle\pm0.3}$ & $49.4{\color{gray}\scriptscriptstyle\pm0.3}$ & $54.2{\color{gray}\scriptscriptstyle\pm0.2}$ \\
 \cmidrule(lr){2-6}
 & CDA + LPLD & $21.9{\color{gray}\scriptscriptstyle\phantom{0.000}}$ & $34.1{\color{gray}\scriptscriptstyle\phantom{0.000}}$ & $46.1{\color{gray}\scriptscriptstyle\phantom{0.000}}$ & $53.3{\color{gray}\scriptscriptstyle\phantom{0.000}}$ \\
 & CDA + Ours & $31.7{\color{gray}\scriptscriptstyle\pm0.3}$ & $42.7{\color{gray}\scriptscriptstyle\pm0.2}$ & $50.5{\color{gray}\scriptscriptstyle\pm0.1}$ & $55.5{\color{gray}\scriptscriptstyle\pm0.3}$ \\
\midrule[1pt]
\multirow{9}{*}{\rotatebox[origin=c]{90}{\textbf{Compression 30×}}}
 & RDED + LPLD & $30.8{\color{gray}\scriptscriptstyle\phantom{0.000}}$ & $39.5{\color{gray}\scriptscriptstyle\phantom{0.000}}$ & $49.7{\color{gray}\scriptscriptstyle\phantom{0.000}}$ & $55.3{\color{gray}\scriptscriptstyle\phantom{0.000}}$ \\
 & RDED + Ours & {\boldmath$35.2{\color{gray}\scriptscriptstyle\pm0.2}$} & {\boldmath$46.4{\color{gray}\scriptscriptstyle\pm0.2}$} & {\boldmath$52.8{\color{gray}\scriptscriptstyle\pm0.2}$} & $56.7{\color{gray}\scriptscriptstyle\pm0.1}$ \\
 \cmidrule(lr){2-6}
 & LPLD + LPLD & $23.1{\color{gray}\scriptscriptstyle\pm0.1}$ & $35.9{\color{gray}\scriptscriptstyle\pm0.3}$ & $48.6{\color{gray}\scriptscriptstyle\pm0.2}$ & $55.2{\color{gray}\scriptscriptstyle\pm0.1}$ \\
 & LPLD + Ours & $32.7{\color{gray}\scriptscriptstyle\pm0.3}$ & $45.0{\color{gray}\scriptscriptstyle\pm0.4}$ & $52.6{\color{gray}\scriptscriptstyle\pm0.3}$ & {\boldmath$56.8{\color{gray}\scriptscriptstyle\pm0.2}$} \\
 \cmidrule(lr){2-6}
 & SRE2L + LPLD & $14.1{\color{gray}\scriptscriptstyle\phantom{0.000}}$ & $24.5{\color{gray}\scriptscriptstyle\phantom{0.000}}$ & $37.2{\color{gray}\scriptscriptstyle\phantom{0.000}}$ & $46.7{\color{gray}\scriptscriptstyle\phantom{0.000}}$ \\
 & SRE2L + Ours & $30.5{\color{gray}\scriptscriptstyle\pm0.22}$ & $40.7{\color{gray}\scriptscriptstyle\pm0.2}$ & $48.0{\color{gray}\scriptscriptstyle\pm0.3}$ & $52.8{\color{gray}\scriptscriptstyle\pm0.1}$ \\
 \cmidrule(lr){2-6}
 & CDA + LPLD & $14.2{\color{gray}\scriptscriptstyle\phantom{0.000}}$ & $27.5{\color{gray}\scriptscriptstyle\phantom{0.000}}$ & $41.8{\color{gray}\scriptscriptstyle\phantom{0.000}}$ & $49.7{\color{gray}\scriptscriptstyle\phantom{0.000}}$ \\
 & CDA + Ours & $29.7{\color{gray}\scriptscriptstyle\pm0.6}$ & $41.8{\color{gray}\scriptscriptstyle\pm0.3}$ & $49.6{\color{gray}\scriptscriptstyle\pm0.1}$ & $54.5{\color{gray}\scriptscriptstyle\pm0.1}$ \\
\midrule[1pt]
\multirow{9}{*}{\rotatebox[origin=c]{90}{\textbf{Compression 40×}}}
 & RDED + LPLD & $29.1{\color{gray}\scriptscriptstyle\phantom{0.000}}$ & $38.4{\color{gray}\scriptscriptstyle\phantom{0.000}}$ & $48.7{\color{gray}\scriptscriptstyle\phantom{0.000}}$ & $54.2{\color{gray}\scriptscriptstyle\phantom{0.000}}$ \\
 & RDED + Ours & {\boldmath$36.3{\color{gray}\scriptscriptstyle\pm0.3}$} & $40.8{\color{gray}\scriptscriptstyle\pm0.7}$ & {\boldmath$51.6{\color{gray}\scriptscriptstyle\pm0.1}$} & {\boldmath$55.2{\color{gray}\scriptscriptstyle\pm0.2}$} \\
 \cmidrule(lr){2-6}
 & LPLD + LPLD & $20.2{\color{gray}\scriptscriptstyle\pm0.3}$ & $33.0{\color{gray}\scriptscriptstyle\pm0.6}$ & $46.7{\color{gray}\scriptscriptstyle\pm0.3}$ & $54.0{\color{gray}\scriptscriptstyle\pm0.8}$ \\
 & LPLD + Ours & $33.7{\color{gray}\scriptscriptstyle\pm0.5}$ & {\boldmath$44.2{\color{gray}\scriptscriptstyle\pm0.5}$} & $51.6{\color{gray}\scriptscriptstyle\pm0.21}$ & $54.8{\color{gray}\scriptscriptstyle\pm0.2}$ \\
 \cmidrule(lr){2-6}
 & SRE2L + LPLD & $11.4{\color{gray}\scriptscriptstyle\phantom{0.000}}$ & $21.7{\color{gray}\scriptscriptstyle\phantom{0.000}}$ & $35.5{\color{gray}\scriptscriptstyle\phantom{0.000}}$ & $44.4{\color{gray}\scriptscriptstyle\phantom{0.000}}$ \\
 & SRE2L + Ours & $28.3{\color{gray}\scriptscriptstyle\pm0.5}$ & $39.3{\color{gray}\scriptscriptstyle\pm0.9}$ & $47.4{\color{gray}\scriptscriptstyle\pm0.6}$ & $51.3{\color{gray}\scriptscriptstyle\pm0.2}$ \\
 \cmidrule(lr){2-6}
 & CDA + LPLD & $13.2{\color{gray}\scriptscriptstyle\phantom{0.000}}$ & $24.0{\color{gray}\scriptscriptstyle\phantom{0.000}}$ & $38.0{\color{gray}\scriptscriptstyle\phantom{0.000}}$ & $47.2{\color{gray}\scriptscriptstyle\phantom{0.000}}$ \\
 & CDA + Ours & $29.4{\color{gray}\scriptscriptstyle\pm0.2}$ & $40.9{\color{gray}\scriptscriptstyle\pm0.4}$ & $48.9{\color{gray}\scriptscriptstyle\pm0.2}$ & $52.6{\color{gray}\scriptscriptstyle\pm0.1}$ \\
\bottomrule
\end{tabular}
}
\caption{
Test accuracy percent on ImageNet-1K across soft label compression ratios and images per class (IPC). Results combine a data synthesis method (SRe2L, CDA, RDED or LPLD) with label compression by LPLD or our VQAE using ResNet18 as both teacher and student and average three runs. Standard deviations appear for our results and no compression numbers only; LPLD stds are blank when no std was reported in their work.
}
\label{tab: main_table}
\end{table}
\flushbottom  

We conduct extensive experiments to measure the effectiveness of our approach in both vision and language tasks under various storage budgets. Our vision experiments were conducted using 12 NVIDIA RTX A6000 GPUs, while the language experiments were run on 4 NVIDIA L40 GPUs.

\subsection{Dataset Distillation - Images}
We evaluate our method on the challenging ImageNet-1K dataset under varying soft-label compression ratios, ranging from \textbf{10$\times$ to 200$\times$}. Each ratio reflects the reduction in the size of information required to transmit soft labels, relative to a baseline where full soft labels are communicated using half-precision (float16) for every training epoch.

\vspace{0.1in}

Our baseline for label compression is LPLD~\cite{xiao2024are}, which reduces storage overhead by discarding the soft labels associated with randomly selected training batches. In contrast, our approach employs a VQAE that encodes soft labels using discrete code indices from a learnable codebook. The compression ratio is governed by three key parameters: the number of code vectors \( k \), their dimensionality \( d_c \), and latent dimension $d_h$. For design simplicity we decided to always set $d_h$ to be divisible by $d_c$. By tuning the aforementioned hyperparameters, we achieved a match to the target communication budget under various evaluation settings, thereby enabling a fair comparison. In cases where precise matching was not possible we slightly reduced our method's parameter budget to ensure fair comparison. For instance using Eq. \ref{eq:comp_ratio} and proper bit conversions, one can achieve a $10\times$ soft label compression by setting $d_h=795$, $d_c=5$, and $k=1024$. Detailed values of these hyperparameters, as well as the procedure for computing the total communication cost across 300 training epochs under various IPC settings, are provided in the supplementary material.


\begin{figure}[!b]
\vspace{-.4in}
\centering
\includegraphics[width=1.\linewidth]{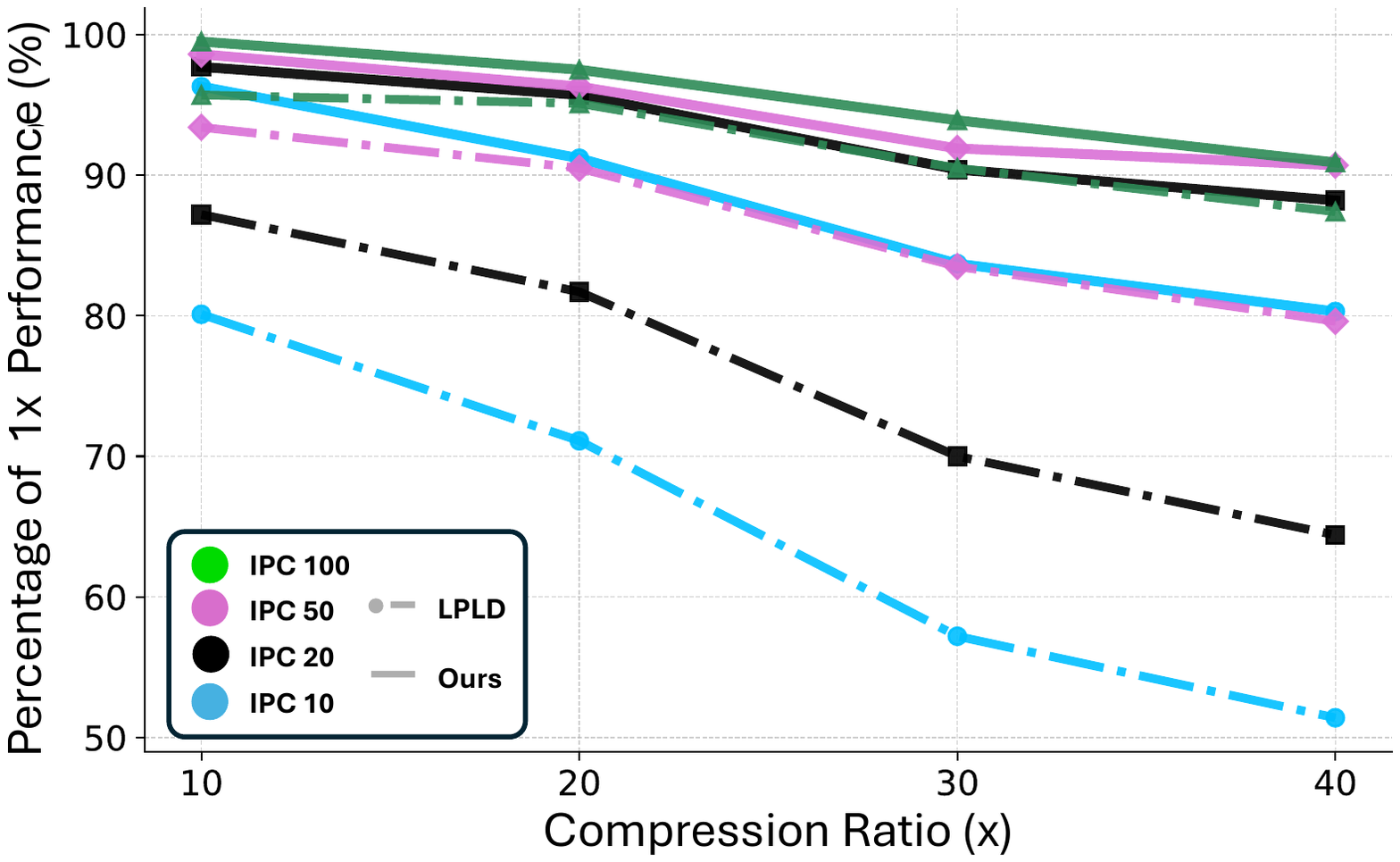}
\caption{Performance ratio of training with compressed over uncompressed soft labels at different compression levels, averaged across image distillation techniques presented in Table \ref{tab: main_table} at each IPC level. Our method consistently outperforms LPLD across all compression factors.}
\label{fig:percentage_performance}
\end{figure}

We gathered the results in Table \ref{tab: main_table}. Our objective is to disentangle the individual contributions of \textit{data synthesis} and \textit{label compression} within the dataset distillation pipeline. To this end, a table entry denoted as \textit{Method A + Method B} indicates that the data is synthesized using Method A, while the soft labels are compressed using Method B (namely ours or LPLD). Our baselines include SRe2L ~\cite{yin2023squeeze}, CDA ~\cite{yin2024dataset}, RDED ~\cite{sun2024diversity}, and LPLD ~\cite{xiao2024are}. Notably, the community has recently proposed a growing number of dataset distillation frameworks, including approaches that leverage diffusion models to synthesize distilled images \cite{su2024d,chen2025influenceguided}. Our contribution is orthogonal to these methods, as it focuses on compressing soft labels and can readily complement any of these frameworks. We believe requiring comparisons with all existing methods, particularly those that are not directly related to our contribution, would be impractical and computationally prohibitive.

We observe consistent performance gains, particularly under higher compression ratios and lower IPCs. To further highlight our contributions, Figure \ref{fig:percentage_performance} illustrates the percentage of maintained accuracy (i.e., the dataset distillation performance ratio with and without soft-label compression) averaged across the dataset distillation techniques presented in Table \ref{tab: main_table} at each IPC level. Our method consistently outperforms LPLD at every evaluated compression ratio, demonstrating superior retention of soft-label information under stringent storage constraints.

In all the experiments, both the teacher and the student have the ResNet18 architecture. For the majority of the experiments, we used an AdamW optimizer with $\text{lr}=0.001$ and $\text{weight\_decay}=0.01$. More detailed hyperparameter tuning, as well as results for compression ratios from 100× to 200×, are included in the supplementary material.

\subsection{Token-Level Soft-Label Compression - LLMs}

\begin{table}[t]
\centering

\resizebox{\columnwidth}{!}{%
\footnotesize
\begin{tabular}{@{}l c l ccc@{}}
\toprule
\textbf{Model} & \textbf{\#Params} & \textbf{Method} &
\textbf{Dolly} & \textbf{SelfInst} & \textbf{Vicuna} \\
\midrule
\multirow{9}{*}{GPT‑2}
& 1.5B & Teacher      & 27.6 & 14.3 & 16.3 \\
\cmidrule(lr){2-6}
& \multirow{4}{*}{120M}
& SFT w/o KD  & 23.3 & 10.0 & 14.7 \\
& & KD         & 22.8 & 10.8 & 13.4 \\
& & SeqKD      & 22.7 & 10.1 & 14.3 \\
& & Ours       & 23.5 & 10.4 & 15.2 \\
\cmidrule(lr){2-6}
& \multirow{4}{*}{760M}
& SFT w/o KD  & 25.4 & 12.4 & 16.1 \\
& & KD         & 25.9 & 13.4 & 16.9 \\
& & SeqKD      & 25.6 & 14.0 & 15.9 \\
& & Ours       & 25.7 & 13.3 & 16.3 \\
\midrule
\multirow{5}{*}{LLaMA}
& 13B & Teacher      & 29.7 & 23.4 & 19.4 \\
\cmidrule(lr){2-6}
& \multirow{4}{*}{7B}
& SFT w/o KD  & 26.3 & 20.8 & 17.5 \\
& & KD         & 27.4 & 20.2 & 18.4 \\
& & SeqKD      & 27.5 & 20.8 & 18.1 \\
& & Ours       & 28.2 & 20.5 & 18.7 \\
\bottomrule
\end{tabular}}
\caption{Comparison of distillation methods on GPT‑2 and LLaMA models.}
\label{tab:llm_results}
\end{table}

With growing awareness of the storage overhead associated with caching soft labels in vision tasks, we argue that the problem is even more severe in the language domain due to the vast output space of large language models (LLMs). In the context of dataset distillation for images, compression typically involves reducing both the input images and their associated soft labels. By contrast, in the language domain, the text itself is highly compressible, so the primary bottleneck arises from storing soft labels. LLMs typically use tokenizers with vocabularies exceeding 50,000 tokens, making it impractical to cache soft labels ahead of token-level distillation. Storing logits over such large vocabularies for billions of training tokens would require petabytes disk space. As a result, prior works \cite{gu2023minillm, muralidharan2024compact} typically resort to online teacher inference during distillation or fall back on sequence-level supervision to reduce storage demands. While performing online inference of the teacher can be acceptable in the traditional knowledge distillation setting, it defeats the main purpose of dataset distillation, which is to create a compact dataset that can be reused without repeatedly querying the teacher.

Table~\ref{tab:llm_results} presents our results on the generation task, following the training and evaluation protocols in \cite{gu2023minillm}. Since, to the best of our knowledge, there is no prior work on compressing soft labels for LLMs, we compare our method against different LLM knowledge distillation approaches as baselines. Results show our logit compression matches or outperforms standard knowledge distillation.

All GPT‑2 models are pretrained on the OpenWebText corpus \cite{Gokaslan2019OpenWeb}, with the teacher further finetuned on the \textit{databricks‑dolly‑15K}\footnote{\url{https://github.com/databrickslabs/dolly/tree/master}} dataset for token‑level distillation. For the LLaMA experiments, models are pretrained on the RoBERTa corpus \cite{liu2019roberta}. Each student model is distilled on the \textit{databricks‑dolly‑15K} dataset and evaluated on the Dolly, Self‑Instruct \cite{wang-etal-2023-self-instruct}, and Vicuna \cite{vicuna2023} benchmarks. Following \cite{gu2023minillm}, we compare against three baselines: supervised fine‑tuning without distillation, vanilla logit‑level knowledge distillation, and sequence‑level knowledge distillation, using the results reported in \cite{gu2023minillm}.

Moreover, unlike the vision experiments, we observe that language tasks benefit significantly from first selecting the top‑$k$ teacher logits (with $k=20$) and then applying vector quantization on these values. In this setting, both the top‑$k$ indices and the VQ‑AE parameters are cached. The final ROUGE‑L scores \cite{lin2004rouge} are reported in Table~\ref{tab:llm_results}.



It is worth considering that the GPT-2 tokenizer has a vocab size of $50,257$ and responses from the Dolly training subset include $\approx1.2$M tokens.\emph{Caching the soft labels in the Vanilla KD setting requires $\approx112$GB of storage. Using our framework, we managed to reduce the memory requirement of storing the teacher soft labels to $200$MB.} \textbf{This accounts for a $560\times$ reduction in required storage,} while saving significant GPU memory without the need for online teacher inference in the distillation phase.

\begin{table}[t]
\centering

\renewcommand{\arraystretch}{1.}
\footnotesize  
\begin{tabular}{l l c}
\toprule
\textbf{Comp.} & \textbf{Method} & \textbf{Value} \\
\midrule
\textbf{1×} & No compression & 39.3 \\
\midrule
\multirow{2}{*}{\textbf{$<$10×}}
& Quant. (7.98×, 2 bits) & 34.1 \\
& Quant. (5.36×, 3 bits) & 36.1 \\
\midrule
\multirow{5}{*}{\textbf{40×}}
& PCA (24 PCs) & 8.9 \\
& RPCA (24 PCs) & 8.1 \\
& Topk (k=15) & 20.1 \\
\cmidrule(lr){2-3}
& Ours w/o AE & 12.4 \\
& Ours & 36.3 \\
\midrule
\multirow{4}{*}{\textbf{100×}}
& PCA (10 PCs) & 2.3 \\
& RPCA (10 PCs) & 2.2 \\
& Topk (k=6) & 18.9 \\
\cmidrule(lr){2-3}
& Ours & 25.4 \\
\bottomrule
\end{tabular}
\caption{Baselines on ImageNet‑1K at several compressions.}
\label{tab:simple_baseline}
\end{table}

\begin{figure}[!b]
\centering
\hspace{-.4in}
\includegraphics[width=1.\linewidth]{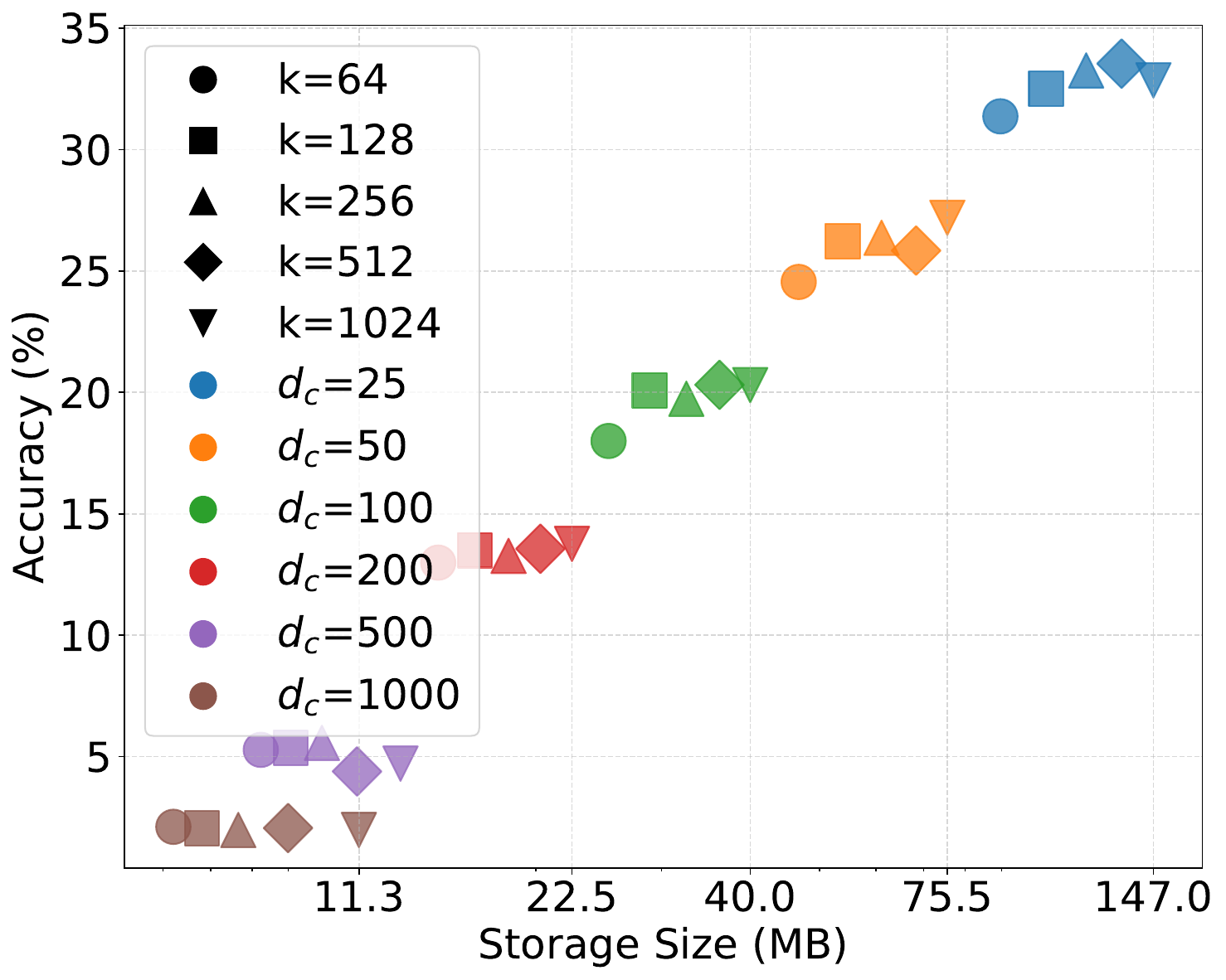}
\caption{Accuracy vs. storage across codebook size \(k\) and block size \(d_c\).}
\label{fig:accuracy_vs_storage}
\end{figure}

\section{Ablation studies}

We conducted extensive ablation studies on the RDED data to assess the impact of different VQ-AE components.
\subsection{Design choices and simple baselines}
Table \ref{tab:simple_baseline} compares our method with the baselines shown below on ImageNet-1K at various compression ratios, using IPC=10 and ResNet-18 as both teacher and student models.

\begin{enumerate}
    \item \textbf{Vanilla Quantization:} Learning $4$ and $8$ quantization levels for the teacher probabilities. 

    \item \textbf{PCA:} Using PCA for compressing the labels. We set the compression ratio to $40\times$ and $100\times$ and calculated the appropriate number of components. 

    \item \textbf{R-PCA:} Similar to the concurrent work of \cite{yuan2025score}, we use robust PCA (R-PCA) for label compression. As their code was not publicly available at the time of submission, we implemented R-PCA within our own framework.

    \item \textbf{Topk:} The topK teacher logit probabilities along with their indices are communicated/stored. 

    \item \textbf{Ours w/o AE:} We discard the encoder and decoder projection matrices and directly use the quantized vector formed from the codes for the compression. 

\end{enumerate}


The results clearly show the superiority of our method with all its components over these baselines.


\subsection{Cross-architectural  Analysis}


We evaluate the effectiveness of our compression method in a cross-architectural setting, where the teacher and student models belong to different families. Results are presented in Table~\ref{tab:compression_archs}. Specifically, we use ResNet-50 \cite{he2016deep}, ShuffleNet-V2 \cite{ma2018shufflenet}, EfficientNet-B0 \cite{tan2019efficientnet}, and Swin Transformer-Tiny \cite{liu2021swin} as ImageNet-1K teachers, and distill to a ResNet-18 student under IPC=10. The results show that our method performs on par with or better than LPLD's soft label pruning across all teacher architectures.

\begin{table}[h]
\centering

\small                     
\setlength{\tabcolsep}{3pt}

\begin{adjustbox}{max width=\columnwidth}
\renewcommand{\arraystretch}{1.4} 
\small     
\begin{tabular}{c|cc|cc|cc|cc}
\toprule
 & \multicolumn{2}{c|}{\textbf{ResNet 50}}
 & \multicolumn{2}{c|}{\textbf{ShuffleNet V2}}
 & \multicolumn{2}{c|}{\textbf{EfficientNet B0}}
 & \multicolumn{2}{c}{\textbf{Swin Tiny}} \\[2pt]
\textbf{Compression} & LPLD & Ours & LPLD & Ours & LPLD & Ours & LPLD & Ours \\
\midrule
\textbf{20×}  & 26.4 & 25.1 & 34.1 & 33.1 & 21.9 & 24.6 & 14.5 & 15.6 \\
\textbf{40×}  & 21.9 & 23.0 & 28.4 & 28.6 & 17.6 & 22.8 & 11.1 & 13.7 \\
\textbf{100×} & 12.3 & 19.7 & 16.9 & 23.1 & 10.1 & 18.7 &  7.3 & 11.2 \\
\bottomrule
\end{tabular}
\end{adjustbox}
\caption{Accuracy values at several compression rates over four architectures.}
\label{tab:compression_archs}
\end{table}

\subsection{Effect of Number and Dimension of Codes}
\label{sec:levelset}


In Fig.~\ref{fig:accuracy_vs_storage}, we analyze the impact of varying the number of codebook entries ($k$) and code dimensionality ($d_c$) on the performance of a ResNet-18 student, under IPC=10 on ImageNet-1K. We vary $k$ from 64 to 1024 and $d_c$ from 50 to 1000, while fixing $d_h = 1000$. The results show that, for a fixed $d_c$, increasing $k$ yields only marginal performance gains, suggesting that similar accuracy can be achieved with fewer codes and reduced memory. Conversely, increasing $d_c$ improves performance more significantly but requires more storage, as each code becomes larger.

Under a similar evaluation setting, we also varied $k$ and $d_c$ while keeping the overall compression ratio constant. At large numbers of training epochs, we observed that the dominant term in the compression cost scales with $\frac{d_c}{\log_2 k}$. By fixing this ratio, we can explore different combinations of $k$ and $d_c$ that yield equivalent compression rates. Table~\ref{tab:levelset} confirms our hypothesis: performance remains largely stable along these level sets, providing flexibility in hyperparameter selection (see supplementary material for details).

\begin{table}[t]
\centering

\small                     
\begin{adjustbox}{max width=\columnwidth}
\begin{tabular}{l c c c c}
\toprule
& $k=2,\,d_c=5$ & $k=4,\,d_c=10$ & $k=16,\,d_c=20$ & $k=256,\,d_c=40$ \\
\midrule
Acc & 24.2 & 27.9 & 26.9 & 27.7 \\
\bottomrule
\end{tabular}
\end{adjustbox}
\caption{Accuracy (\%) for four $(k,d_c)$ settings at a compression ratio of $75\times$.}
\label{tab:levelset}
\end{table}



\section{Conclusion}
We introduced a simple yet effective vector-quantized autoencoder to compress soft labels in dataset distillation, addressing the often-overlooked storage bottleneck of soft labels. Our approach carefully tackles this critical problem by decoupling label compression from data synthesis, integrating seamlessly with existing pipelines, and achieving strong results across vision and language tasks even under extreme compression. These findings underscore the central role of soft labels and the importance of efficient compression for scalable distillation.

\bibliographystyle{ieeenat_fullname}
\bibliography{main}
\clearpage
\onecolumn


\renewcommand{\thetable}{S\arabic{table}}
\renewcommand{\thefigure}{S\arabic{figure}}
\renewcommand{\theequation}{S\arabic{equation}}

\let\OLDthebibliography\thebibliography
\renewcommand\thebibliography[1]{
  \OLDthebibliography{#1}
  \setlength{\parskip}{0pt}
  \setlength{\itemsep}{0pt plus 0.3ex}
}


\newcommand{\aj}{AJ}                   
\newcommand{\actaa}{Acta Astron.}      
\newcommand{\araa}{ARA\&A}             
\newcommand{\apj}{ApJ}                 
\newcommand{\apjl}{ApJ}                
\newcommand{\apjs}{ApJS}               
\newcommand{\ao}{Appl.~Opt.}           
\newcommand{\apss}{Ap\&SS}             
\newcommand{\aap}{A\&A}                
\newcommand{\aapr}{A\&A~Rev.}          
\newcommand{\aaps}{A\&AS}              
\newcommand{\azh}{AZh}                 
\newcommand{\baas}{BAAS}               
\newcommand{\bac}{Bull. astr. Inst. Czechosl.}
\newcommand{\caa}{Chinese Astron. Astrophys.}
\newcommand{\cjaa}{Chinese J. Astron. Astrophys.}
\newcommand{\icarus}{Icarus}           
\newcommand{\jcap}{J. Cosmology Astropart. Phys.}
\newcommand{\jrasc}{JRASC}             
\newcommand{\memras}{MmRAS}            
\newcommand{\mnras}{MNRAS}             
\newcommand{\na}{New A}                
\newcommand{\nar}{New A Rev.}          
\newcommand{\pra}{Phys.~Rev.~A}        
\newcommand{\prb}{Phys.~Rev.~B}        
\newcommand{\prc}{Phys.~Rev.~C}        
\newcommand{\prd}{Phys.~Rev.~D}        
\newcommand{\pre}{Phys.~Rev.~E}        
\newcommand{\prl}{Phys.~Rev.~Lett.}    
\newcommand{\pasa}{PASA}               
\newcommand{\pasp}{PASP}               
\newcommand{\pasj}{PASJ}               
\newcommand{\rmxaa}{Rev. Mexicana Astron. Astrofis.}%
\newcommand{\qjras}{QJRAS}             
\newcommand{\skytel}{S\&T}             
\newcommand{\solphys}{Sol.~Phys.}      
\newcommand{\sovast}{Soviet~Ast.}      
\newcommand{\ssr}{Space~Sci.~Rev.}     
\newcommand{\zap}{ZAp}                 
\newcommand{\nat}{Nature}              
\newcommand{\iaucirc}{IAU~Circ.}       
\newcommand{\aplett}{Astrophys.~Lett.} 
\newcommand{\apspr}{Astrophys.~Space~Phys.~Res.}
\newcommand{\bain}{Bull.~Astron.~Inst.~Netherlands} 
\newcommand{\fcp}{Fund.~Cosmic~Phys.}  
\newcommand{\gca}{Geochim.~Cosmochim.~Acta}   
\newcommand{\grl}{Geophys.~Res.~Lett.} 
\newcommand{\jcp}{J.~Chem.~Phys.}      
\newcommand{\jgr}{J.~Geophys.~Res.}    
\newcommand{\jqsrt}{J.~Quant.~Spec.~Radiat.~Transf.}
\newcommand{\memsai}{Mem.~Soc.~Astron.~Italiana}
\newcommand{\nphysa}{Nucl.~Phys.~A}   
\newcommand{\physrep}{Phys.~Rep.}   
\newcommand{\physscr}{Phys.~Scr}   
\newcommand{\planss}{Planet.~Space~Sci.}   
\newcommand{\procspie}{Proc.~SPIE}   

\begin{table}[b!]
\small
\centering
\begin{tabular}{c|cccc}
\hline
\textbf{Method} & \textbf{IPC} & \textbf{Image Size (GB)} & \textbf{Label Size (GB)} & \textbf{Total Size (GB)} \\
\hline
\multirow{4}{*}{\rotatebox[origin=c]{90}{RDED}} 
& 10  & 0.127 & 5.588  & 5.715 \\
& 20  & 0.250 & 11.176 & 11.426 \\
& 50  & 0.610 & 27.940 & 28.550 \\
& 100 & 1.230 & 55.879 & 57.109 \\
\hline
\multirow{4}{*}{\rotatebox[origin=c]{90}{SRe2L}} 
& 10  & 0.150 & 5.588  & 5.738 \\
& 20  & 0.300 & 11.176 & 11.476 \\
& 50  & 0.740 & 27.940 & 28.680 \\
& 100 & 1.490 & 55.879 & 57.369 \\
\hline
\multirow{4}{*}{\rotatebox[origin=c]{90}{CDA}} 
& 10  & 0.140 & 5.588  & 5.728 \\
& 20  & 0.290 & 11.176 & 11.466 \\
& 50  & 0.718 & 27.940 & 28.658 \\
& 100 & 1.430 & 55.879 & 57.309 \\
\hline
\multirow{4}{*}{\rotatebox[origin=c]{90}{LPLD}} 
& 10  & 0.160 & 5.588  & 5.748 \\
& 20  & 0.320 & 11.176 & 11.496 \\
& 50  & 0.800 & 27.940 & 28.740 \\
& 100 & 1.590 & 55.879 & 57.469 \\
\hline
\end{tabular}
\caption{Storage breakdown of soft labels and data across methods without any compression}
\label{tab:storage_1x}
\end{table}

\section{Experimental Details}
This section provides a detailed breakdown of the compression rates and the storage costs for both images and soft labels across different methods and IPC values. Moreover, additional results showing the effectiveness of our method for higher IPCs are also provided. 

\subsection{Compression Rate Calculation}
The soft label size for the standard knowledge distillation-based dataset distillation without any compression is computed as follows:

\begin{equation}
    S_{\text{soft}} = \frac{IPC \times C \times C \times \text{EPOCHS} \times 2 ~ \text{Byte}}{1024^2} ~~~ \text{(Megabytes)}
    \label{eq:size_formula}
\end{equation}

\vspace{10mm}

\noindent Here, $IPC$ denotes the number of images per class, and $C$ is the number of classes. We multiply by $2~\text{Bytes}$ to account for the storage cost of half-precision tensors. Based on Eq.~\ref{eq:size_formula}, the image and soft label sizes under various experimental settings without any label compression are summarized in Table~\ref{tab:storage_1x}. Note that $\text{EPOCHS} = 300$.

\subsection{Vector Quantization Hyperparameters}
We provide the hyperparameter details of our method and compare the final storage size with those achieved by LPLD. The storage cost of our method is computed using the following formulation:


\begin{equation}
S_{\text{soft}}^{\text{VQ}} = \frac{
    \overbrace{IPC \times C \times \text{EPOCHS} \times \frac{d_h}{d_c} \times \frac{\log_2(k)}{8 ~ \text{bits}}}^{\text{Batch Data}} 
    + \overbrace{C \times d_h \times 4 ~ \text{Bytes}}^{\text{Decoder}} 
    + \overbrace{k \times d_c \times 4 ~ \text{Bytes}}^{\text{Codebook}} 
}{1024^2} 
\label{eq:vq_formula}
\end{equation}

\begin{table}[b!]
\small
\centering
\begin{tabular}{c|c|c|c}
\hline
\textbf{Rate} & \textbf{IPC} & \textbf{Ours (GB)} & \textbf{LPLD (GB)} \\
\hline
\multirow{4}{*}{10x} 
& 10  & 0.558 & 0.559 \\
& 20  & 1.114 & 1.118 \\
& 50  & 2.779 & 2.794 \\
& 100 & 5.556 & 5.588 \\
\hline
\multirow{4}{*}{20x} 
& 10  & 0.257 & 0.279 \\
& 20  & 0.511 & 0.559 \\
& 50  & 1.272 & 1.397 \\
& 100 & 2.539 & 2.794 \\
\hline
\multirow{4}{*}{30x} 
& 10  & 0.178 & 0.186 \\
& 20  & 0.353 & 0.373 \\
& 50  & 0.877 & 0.931 \\
& 100 & 1.750 & 1.863 \\
\hline
\multirow{4}{*}{40x} 
& 10  & 0.130 & 0.140 \\
& 20  & 0.255 & 0.279 \\
& 50  & 0.632 & 0.698 \\
& 100 & 1.261 & 1.397 \\
\hline
\multirow{2}{*}{100x} 
& 10  & 0.053 & 0.056 \\
& 20  & 0.102 & 0.112 \\

\hline
\multirow{2}{*}{200x} 
& 10  & 0.025 & 0.028 \\
& 20  & 0.046 & 0.056 \\

\hline
\end{tabular}
\caption{Comparison of storage size (GB) between our method and LPLD across IPC levels and compression rates}
\label{tab:soft_comp}
\end{table}

\noindent
Here, $d_c$ and $d_h$ represent the dimensionalities of the codebook vectors and latent vectors, respectively, and $k$ is the number of codes. The division by $8$ in Eq.~\ref{eq:vq_formula} accounts for converting bits to bytes. \\

\noindent Unlike LPLD, where the storage cost is scaled down by a fixed compression or pruning factor, our method requires careful tuning of the hyperparameters $k$, $d_c$, and $d_h$. We use full-precision storage for both the decoder and the codebook, as their contribution to the total cost becomes negligible at higher epoch counts. \\

\noindent The soft label storage costs of our method and LPLD are compared in Table~\ref{tab:soft_comp}, where we target achieving storage sizes equal to or slightly below those of LPLD. The corresponding vector quantization hyperparameters used to obtain these compression rates are listed in Table~\ref{tab:compression_hparams}. Please note that all values assume $\text{EPOCHS} = 300$.

\vspace{5mm}

\begin{table}[ht!]
\small
\centering
\begin{tabular}{c|c|c|c}
\hline
\textbf{Comp. Rate} & \textbf{$d_h$} & \textbf{$d_c$} & \textbf{$k$} \\
\hline
10x  & 795  & 5   & 1024 \\
20x  & 990  & 15  & 2048 \\
30x  & 1000 & 20  & 1024 \\
40x  & 1000 & 25  & 512  \\
100x & 1000 & 50  & 128  \\
200x & 1000 & 100 & 64   \\
\hline
\end{tabular}
\caption{Hyperparameter settings for our method under various compression rates}
\label{tab:compression_hparams}
\end{table}

\begin{table}[b]
\small
\centering
\begin{tabular}{lcc@{\hskip 1cm}cc}
\toprule
\textbf{Method}
  & \multicolumn{2}{c}{\textbf{100×}}
  & \multicolumn{2}{c}{\textbf{200×}} \\
\cmidrule(lr){2-3}\cmidrule(lr){4-5}
  & \textbf{IPC 10} & \textbf{IPC 20}
  & \textbf{IPC 10} & \textbf{IPC 20} \\
\midrule
RDED + LPLD   & $14.09$ & $24.80$   & $9.34$  & $15.26$ \\
RDED + Ours   & $19.07$ & $26.16$ & $\textbf{14.85}$ & $\textbf{24.73}$    \\
\midrule
LPLD + LPLD   & $9.14$  & $17.56$   & $5.11$  & $8.74$  \\
LPLD + Ours   & $\textbf{25.43}$ & $\textbf{33.58}$   & $14.00$ & $22.17$ \\
\midrule
SRE2L + LPLD  & $6.90$  & $15.37$   & $3.51$  & $7.45$  \\
SRE2L + Ours  & $20.96$ & $29.15$   & $12.49$ & $17.21$ \\
\midrule
CDA + LPLD    & $6.02$  & $13.83$   & $2.34$  & $4.91$  \\
CDA + Ours    & $21.63$ & $30.29$   & $12.09$ & $19.37$ \\
\bottomrule
\end{tabular}
\caption{Results for compression ratio of 100× and 200×}
\label{tab:high_ratio}
\end{table}

\subsection{Hyperparameters at a Fixed Compression Rate}
In Section~5.3 of our ablation studies, we investigated whether it is possible to maintain a fixed compression rate while varying the number of codes ($k$) and their dimensionality ($d_c$). To express the compression ratio as a function of $k$ and $d_c$, one can directly apply Eq.~\ref{eq:size_formula} and Eq.~\ref{eq:vq_formula}:

\begin{equation}
    \frac{S_{\text{soft}}}{S_{\text{soft}}^{\text{VQ}}}  = \frac{IPC \times C \times C \times \text{EPOCHS} \times 2}{
    IPC \times C \times \text{EPOCHS} \times \frac{d_h}{d_c} \times \frac{\log_2(k)}{8} + C \times d_h \times 4 + k \times d_c \times 4 } ~~ \text{(Megabytes)}
    \label{eq:hyp_comp_ratio}
\end{equation}

\vspace{5mm}

\noindent As $\text{EPOCHS} \to \infty$, the first term in the denominator dominates the remaining two. Assuming $d_h$ remains constant, Eq.~\ref{eq:hyp_comp_ratio} can be approximated by:

\begin{equation}
    \frac{S_{\text{soft}}}{S_{\text{soft}}^{\text{VQ}}} \approx \frac{d_c}{log_2(k)}
    \label{eq:comp_approx}
\end{equation}

\noindent
Eq.~\ref{eq:comp_approx} illustrates the approximate relationship between $k$ and $d_c$ required to maintain a constant compression rate. In our experiments, we began with $d_c = 5$ and incrementally doubled it up to $d_c = 40$, while squaring the value of $k$ at each step to keep the compression ratio approximately constant at $\sim 75\times$.

\subsection{Simple Baselines Compression}
In Section~5.1 of our ablation studies, we compared our method against PCA, R-PCA, Top-$K$, and quantization. Below, we provide the corresponding calculations used to select each baseline’s hyperparameters. \\

\noindent TopK storage:

\begin{equation}
     S_{\text{soft}}^{\text{TopK}} = \frac{IPC \times C \times \text{EPOCHS} \times (\overbrace{2k}^{\text{Class Prob.}} +   \overbrace{\frac{log_2(C)}{8} \times k}^{\text{TopK Indices}})}{1024^2} 
    \label{eq:topk_size}
\end{equation}
\vspace{5mm}
\noindent Where $k$ is the hyperparameter of top-k sorting. \\

\noindent PCA and R-PCA storage cost: \\ 

\begin{equation}
     S_{\text{soft}}^{\text{PCA}} = \frac{IPC \times C \times \text{EPOCHS} \times \overbrace{2k}^{\text{PC Projections}} +   \overbrace{k \times C \times 2}^{\text{PC Vectors}}}{1024^2} 
    \label{eq:PCA_size}
\end{equation}

\vspace{5mm}

\noindent Where $k$ is the number of the principal components. \\

\noindent For simple quantization, we only learn $k$ number of quantization levels for the probabilities: \\

\begin{equation}
    S_{\text{soft}}^{\text{Quant}} = \frac{IPC \times C \times \text{EPOCHS} \times C \times \overbrace{\frac{log_2(k)}{8}}^{\text{lvl index}} +   \overbrace{2k}^{\text{Quant. lvls}}}{1024^2} 
    \label{eq:quant_size}
\end{equation}

\vspace{5mm}

\noindent In Sec. 5.1., 2-bit and 3-bit quantization corresponds to 4 and 8 levels of quantization. 

\subsection{Extreme Compression Rate}
The results for $100\times$ and $200\times$ compression rates are presented in Table~\ref{tab:high_ratio}. Experiments are conducted on ImageNet-1K with IPC = 10 and IPC = 20. As shown in Table~\ref{tab:high_ratio}, our compression method consistently outperforms LPLD by a significant margin, particularly under higher compression and lower data regimes.


\end{document}